\documentclass[letterpaper, 10 pt, conference]{ieeeconf}  % Comment this line out if you need a4paper

\IEEEoverridecommandlockouts                              % This command is only needed if 
\usepackage{graphics} % for pdf, bitmapped graphics files
\usepackage{epsfig} % for postscript graphics files
\usepackage{mathptmx} % assumes new font selection scheme installed
\usepackage{times} % assumes new font selection scheme installed
\usepackage{amsmath} % assumes amsmath package installed
\usepackage{amssymb}  % assumes amsmath package installed
\usepackage{enumerate}
\usepackage{subfigure}

\title{\LARGE \bf
Estimating Risk Levels of Driving Scenarios through Analysis of Driving Styles for Autonomous Vehicles
}

\author{Songlin Xu$^{1}$ and Jiacheng Zhu$^{*}$% <-this % stops a space
\thanks{$^{1}$Songlin Xu is with the Department of Automation, University of Science and Technology of China, Hefei, 230026, China.
        {\tt\ xsl314@mail.ustc.edu.cn}}%
\thanks{$^{*}$Jiacheng Zhu is with the Department of Mechanical Engineering, Carnegie Mellon University, Pittsburgh, PA 15213, USA.
        {\tt\ jzhu4@andrew.cmu.edu}}%
}

\begin{document}

\maketitle
\thispagestyle{empty}
\pagestyle{empty}

%%%%%%%%%%%%%%%%%%%%%%%%%%%%%%%%%%%%%%%%%%%%%%%%%%%%%%%%%%%%%%%%%%%%%%%%%%%%%%%%
\begin{abstract}
In order to operate safely on the road, autonomous vehicles need not only to be able to 
identify objects in front of them, but also to be able to estimate the risk level of the 
object in front of the vehicle automatically. It is obvious that different objects have different levels of danger to autonomous vehicles. 
An evaluation system is needed to automatically determine the danger level of the object for the autonomous vehicle. It would be too subjective and incomplete if the system were completely 
defined by humans. Based on this, we propose a framework based on  nonparametric Bayesian learning method -- a sticky hierarchical Dirichlet process hidden Markov model(sticky HDP-HMM), and discover the relationship between driving scenarios and driving styles. We use the analysis of driving styles of 
autonomous vehicles to reflect the risk levels of driving scenarios to the vehicles. In this framework, we firstly use sticky HDP-HMM to extract driving styles from the dataset and get different clusters, then an evaluation system is proposed to evaluate and rank the urgency levels of the clusters. Finally, we map the driving scenarios to the ranking results and thus get clusters of driving scenarios in
different risk levels. More importantly, we find the relationship between driving scenarios and driving
styles. The experiment shows that our framework can cluster and rank driving styles of different urgency
levels and find the relationship between driving scenarios and driving styles and the conclusions also 
fit people's common sense when driving. Furthermore, this framework can be used for autonomous vehicles to estimate risk levels of driving scenarios and help them make precise and safe decisions.
\end{abstract}

%%%%%%%%%%%%%%%%%%%%%%%%%%%%%%%%%%%%%%%%%%%%%%%%%%%%%%%%%%%%%%%%%%%%%%%%%%%%%%%%

\section{INTRODUCTION}
\label{section: introduction}
Safety plays a vital role in autonomous vehicles, whether for the purpose of protecting 
pedestrians or the vehicles. 

With the development of computer vision, lots of 
methods such as Fast R-CNN \cite{DBLP:journals/corr/Girshick15} can be used to detect objects which improves the autonomy of the  vehicles and also the safety of themselves. However, being able to 
detect the object in front of the vehicle with great precision is not a guarantee of complete safety, 
because detecting an object does not mean that the vehicle knows about the danger of the object to the
vehicle itself. Hence, keeping autonomous vehicles as safe as possible is still a problem.

Since driving styles have a lot to do with vehicles' safety, lots of research has focused on evaluation, classification and recognition of driving styles. There're also various
ways to classify and recognize driving styles, whether using inertial sensors\cite{6629603} or a smart phone\cite{6083078}. Brombacher et al.\cite{7915497} used artificial neural networks to detect driving event and classify driving styles,
whereas Murphey et al.\cite{4938719} used jerk analysis for the driver's style classification. Wang et al.\cite{8015191} implemented 
a semisupervised support vector machine to classify driving styles and reduce the labeling effort at the
same time. Driving styles can be extracted by introducing thresholds for velocity or acceleration.
Wang et al.\cite{DBLP:journals/corr/abs-1708-08986} used threshold to describe driving styles and used primitive driving patterns with bayesian 
nonparametric approaches to analyze driving styles. And driving styles can also be learned from 
demonstration\cite{7139555} and experts\cite{Silver2012LearningAD}. 

But even if we evaluate and classify driving styles properly, how to define the risk levels is still
a challenge for autonomous vehicles. Eboli et al.\cite{EBOLI201729} combined objective and subjective measures of 
driving styles to define the accident risk level. Vaitkus et al.\cite{6957429} proposed pattern recognition approach to classify
driving styles into aggressive or normal patterns automatically using accelerometer data when driving the same route
in different driving styles. Siordia et al.\cite{5548130} classified driving risk based on experts evaluation. However, estimating risk levels by humans can be subjective and sometimes 
incomplete. 

Based on this, we use a sticky hierarchical Dirichlet process hidden Markov model to 
extract driving styles from traffic data, which avoids the subjectivity and is more complete. The sticky
HDP-HMM method is a nonparametric Bayesian learning method which can extract features without prior 
knowledge. \cite{DBLP:journals/corr/abs-1805-07643} used HDP-HSMM to cluster driving data to evaluate energy efficiency. In this work, we propose an evaluation system to rank the cluster results of sticky HDP-HMM from driving data and use this to estimate risk levels of the objects. Lots of research focused on the classification of different driving styles \cite{7915497} \cite{4938719}
but didn't find the relationship between driving scenarios and driving styles. \cite{5548130} considered driving risk classification but it also ignored the interaction between the ego cars and driving scenarios. Hence, we propose a framework combining driving scenarios
and driving levels to find the relationship between them and evaluate risk levels of driving scenarios
through the analysis of driving styles based on nonparametric Bayesian learning.

%%%%%%%%%%%%%%%%%%%%%%%%%%%%%%%%%%
\begin{figure*}
\centering
\includegraphics[width=7in]{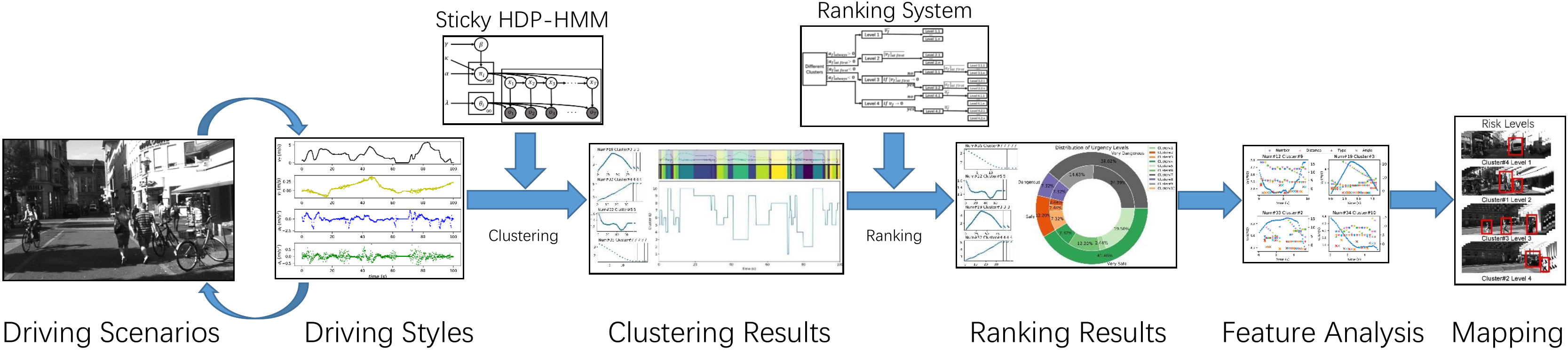}
\caption{Our framework. We firstly use sticky HDP-HMM to extract driving styles from time series driving data 
and then rank the clusters according to the urgency levels. Then we analyze the features of driving scenarios and find the relationship between driving styles and driving scenarios. Finally we map the driving scenarios to the ranking results and get the ranked driving scenarios according to their risk levels.}
\label{fig: framework}
\end{figure*}
%%%%%%%%%%%%%%%%%%%%%%%%%%%%%%%%%%
\begin{figure}
\centering
\subfigure[Hidden Markov Model]{
\includegraphics[width=2.5in]{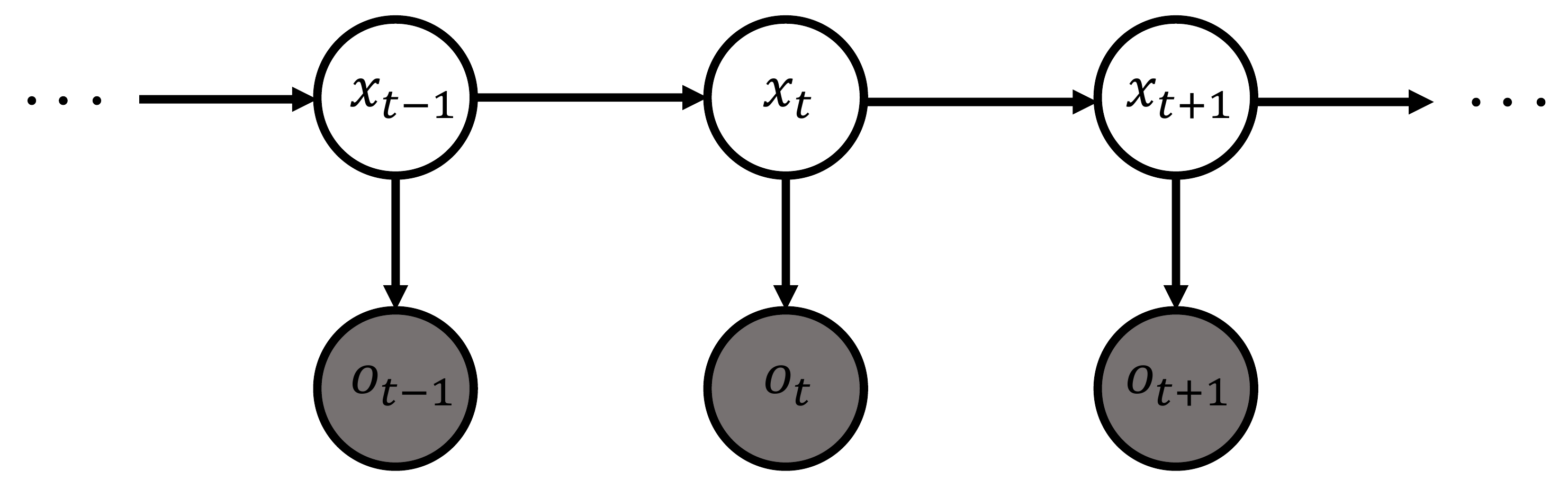}
}
\quad
\subfigure[the sticky HDP-HMM]{
\includegraphics[width=2.5in]{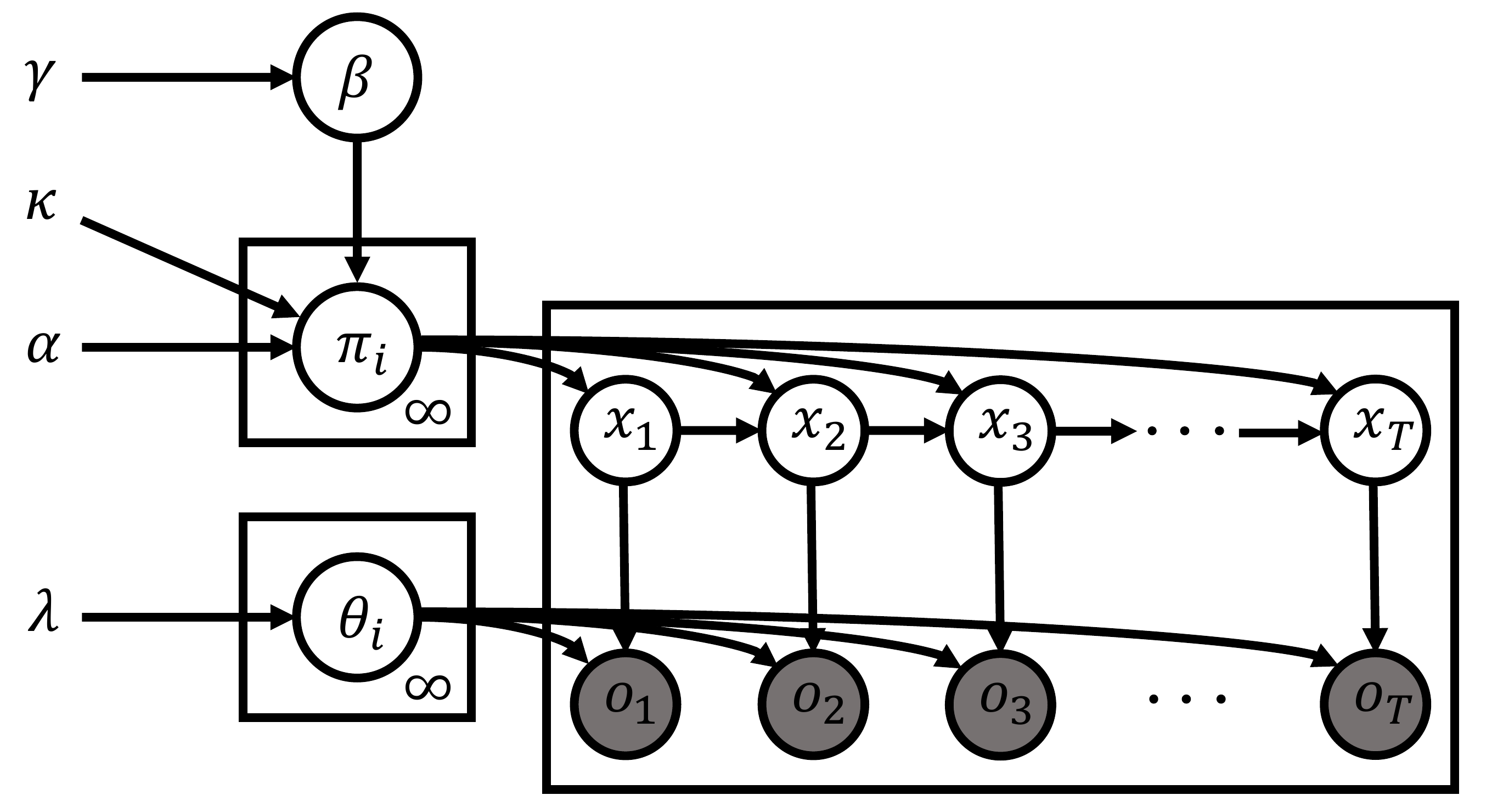}
}
\caption{The model which is used to cluster driving data.}
\label{fig: HDP-HMM}
\end{figure}
%%%%%%%%%%%%%%%%%%%%%%%%%%%%%%%%%%%%
The paper presents the contributions as follows.
\begin{itemize}

\item Presenting a framework based on nonparametric Bayesian learning method to extract 
driving styles. 
\item Proposing an evaluation system to evaluate risk levels of driving scenarios through the analysis of driving styles.
\item Building a bridge between driving scenarios and driving styles and discovering the relationship between them which also fits people's common sense.

\end{itemize}

The remainder of this paper is organized as follows. Section \ref{section: methods} describes the proposed framework along with
the nonparametric Bayesian learning method and ranking method. Section \ref{section: experiment and analysis} introduces the dataset we use and the experiment procedure. Experiment results and analysis are also presented in this section. Finally, the conclusions and future work are given in Section \ref{section: conclusion and future work}.
\section{METHODS}
\label{section: methods} 
In this section we will describe our framework as well as the nonparametric Bayesian learning method and 
the ranking algorithm. We use this framework to extract driving styles from time series driving data 
and rank the clusters according to the urgency levels and finally map driving scenarios to the ranking
results where we find the relationship between driving styles and driving scenarios which also fit 
people's common sense of driving.

\subsection{Framework} As we know the driving scenarios and the ego 
car will affect each other. Besides, since driving scenarios will have an influence on the ego car,
we can infer the risk levels of driving scenarios from analysis of driving styles. For instance, when 
a car is in a dangerous scenario, the driver will brake and the car will slow down. So from the 
reduction of velocity we can infer that the car is in a dangerous situation. And this is the original
source of our framework. There're four main steps in this framework which is shown in Fig.\ref{fig: framework}:  

\begin{enumerate}[(1)]
\item Clustering driving styles through a sticky HDP-HMM method.
\item Ranking the clusters according to the urgency levels of different clusters.
\item Analyzing the influence of driving scenarios' features on the driving styles and getting the 
relationship between driving scenarios and driving styles.
\item Mapping the results of ranking driving styles clusters to driving scenarios and inferring the
risk levels of driving scenarios to the ego car. 
\end{enumerate}

We'll introduce our clustering and ranking method below and the analyzing and mapping results will also
be shown in our experiment part.
\subsection{Clustering method} This part will introduce the sticky hierarchical Dirichlet process hidden Markov model(sticky HDP-HMM). We'll first introduce
hidden Markov model(HMM) and hierarchical Dirichlet process(HDP) and then detail the sticky HDP-HMM
method.

\subsubsection{Hidden Markov Model(HMM)}
Hidden Markov model is a statistical model that can be used to describe a markov process with hidden
unknown parameters, which can be determined from observable parameters. Thus, the HMM is composed of
two layers: a hidden layer and an observation layer. HMM is often used to solve mathematical problems with
implicit conditions. We first assume the observed result is $O = o_1,o_2,...,o_T$ and the hidden condition
is $X = x_1,x_2,...,x_T$, which are both time series data. Then the probability of occurrence of event $O$
is described by the following formula:
\begin{equation}
\label{equation: (1)}
P(O) = \sum\limits_{X} P(O|X)P(X)
\end{equation}
We denote the transition probability from $x_i$ to $x_j$ as $\pi_{ij}$, 
thus $\pi_{ij} = P(x_{t+1}=j|x_t=i)$ and $\pi_i = \pi_{i1} + \pi_{i2} + \pi_{i3} + \cdots$.
So we get:
\begin{equation}
\label{equation: (2)}
x_{t+1}|x_t \sim \pi_{x_t}
\end{equation}
Then we denote the emission probability from $x_t$ to $o_t$ as $P(o_t|x_t)$ and 
$P(o_t|x_t) = P(o_t|x_1,...,x_{t-1},x_t,o_1,...,o_{t-1})$.
And we get:
\begin{equation}
\label{equation: (3)}
o_t|x_t, \theta_{x_t} \sim F(\theta_{x_t})
\end{equation}
$F(\theta_{x_t})$ is the emission function which describes the way from $x_t$ to $o_t$ and $\theta_{x_t}$
is called emission parameter\cite{DBLP:journals/corr/abs-1709-03553}.

\subsubsection{Hierarchical Dirichlet Process(HDP)}
The Dirichlet process, denoted by $DP(\gamma,H)$, is parameterized by a base distribution $H$ and a real
number $\gamma$ which is a positive value. And $\gamma$ describes how discrete the model is.
The Dirichlet process can also be viewed as a stick-breaking process which is described as follows:
\begin{equation}
\label{equation: (4)}
G_0 = \sum\limits_{i=1}^{\infty} \beta_i \delta_{\theta_i}
\end{equation}
The $\theta_i$ are distributed according to $H$ ($\theta \sim H$) which is a base distribution as 
mentioned before and 
$\delta_{\theta_i}$ is an indicator function. And the $\beta_i$ are given by a so-called stick-breaking
process formulated as follows:
\begin{equation}
\label{equation: (5)}
\beta_i = \nu_i \prod_{l=1}^{i-1} (1-\nu_l) \quad i = 1,2,3, \cdots
\end{equation}
$\nu_i$ are independent random variables with the beta distribution($\nu_i \sim Beta(1,\gamma)$).

The hierarchical Dirichlet process(HDP) is a model that can cluster grouped data using Dirichlet process
discussed above. More specifically, it uses the Dirichlet process to share the basic distribution for 
each group of data. Assuming each random probability measure $G_j$ has distribution given by a 
Dirichlet process, thus we have:
\begin{equation}
\label{equation: (6)}
G_j|G_0 \sim DP(\alpha,G_0)
\end{equation}
In the formula, $\alpha$ is the concentration parameter and $G_0$ is the base distribution and from 
formula (\ref{equation: (4)}) we can obtain that:
\begin{equation}
\label{equation: (7)}
G_0 \sim DP(\gamma,H)
\end{equation}
Finally, from (\ref{equation: (4)})(\ref{equation: (5)})(\ref{equation: (6)})(\ref{equation: (7)}) we can get that:
\begin{gather}
\label{equation: 8a}
G_j = \sum\limits_{i=1}^{\infty} \pi_{ji} \delta_{\theta_i}  \tag{8a}\\
\label{equation: 8b}
\pi_j|\alpha,\beta \sim DP(\alpha,\beta)  \tag{8b}\\
\label{equation: 8c}
\theta_i|H \sim H  \tag{8c}\\
\notag
\end{gather}
\subsubsection{Sticky HDP-HMM}
In the sticky HDP-HMM method, a positive value $\kappa$ is added to increase the expected probability of
self-transition \cite{Fox2009The} and compared with (8b) we obtain:
\begin{equation}
\label{equation: (9)}
\pi_i|\alpha,\beta,\kappa \sim DP(\alpha+\kappa,\frac{\alpha\beta+\kappa\delta_i}{\alpha+\kappa})
\tag{9}
\end{equation}
Combining all of these formulas above, we can obtain that:
\begin{gather}
\label{equation: 10a}
x_{t+1}|x_t \sim \pi_{x_t}  \quad t=1,2,3, \cdots ,T       \tag{10a}\\
\label{equation: 10b}
o_t|x_t, \theta_{x_t} \sim F(\theta_{x_t})  \quad t=1,2,3, \cdots ,T       \tag{10b}\\
\label{equation: 10c}
\theta_i|H \sim H    \quad i=1,2,3, \cdots     \tag{10c}\\
\label{equation: 10d}
\pi_i|\alpha,\beta,\kappa \sim DP(\alpha+\kappa,\frac{\alpha\beta+\kappa\delta_i}{\alpha+\kappa}) 
\quad i=1,2,3, \cdots  
\tag{10d}\\
\notag
\end{gather}
We use Gaussian emissions as the emission function $F(\theta_i)$ to determine the observation model
and the $\theta_i$ is set as $\theta_i=[\mu_{p_i},\sigma_{p_i}]$. In addition, we treat the ego car's
dynamic driving process as a combination of different driving styles and this dynamic process can be
modeled as the HMM. Then the HDP is used to develop the HMM with an infinite state space\cite{NIPS2004_2698} to learn
from data. Additionally, a positive value $\kappa$ is added to increase the expected probability of
self-transition. Finally, we obtain the sticky HDP-HMM model to process the driving data.
\subsection{Ranking method}

\begin{figure}
\centering
\includegraphics[width=3.5in]{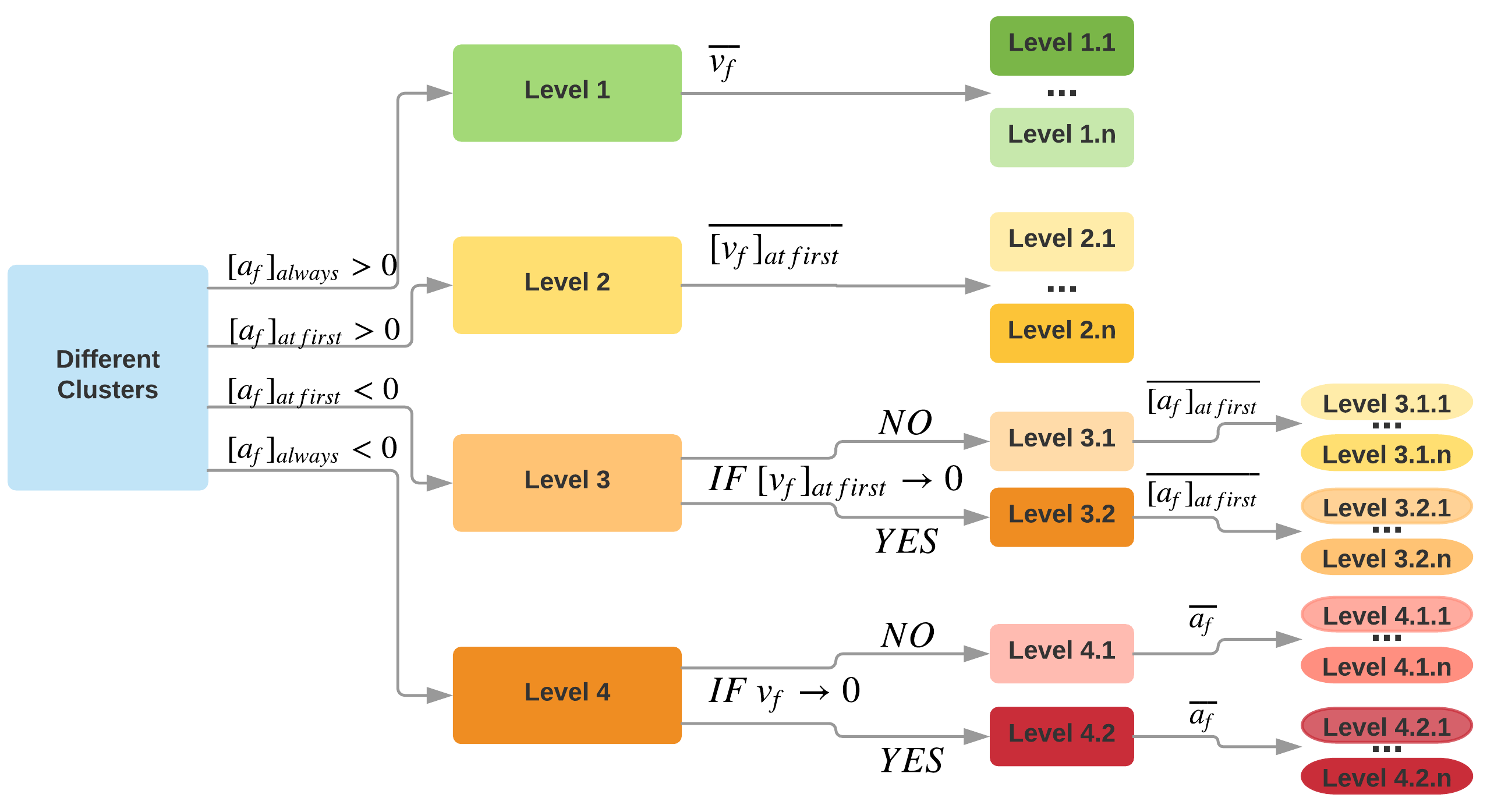}
\caption{Ranking process. We first divide the driving styles into four levels based on whether
the acceleration is always greater/less than zero. Then we analyze each level respectively.}
\label{fig: Ranking process}
\end{figure}

Tanishita et al.\cite{TANISHITA2017107} showed that not only the mean speeds but also changes in the mean speeds affected the per vehicle-kilometer traffic accident rates. However, they arbitrarily distinguished six areas of different speed levels which can be incomplete and subjective. 
\cite{AFWAHLBERG200483} studied using driver acceleration behavior as an accident predictor. However, only using acceleration
is incomplete and there is no very significant correlation between the acceleration variables and accidents found.
Thus, instead of dividing driving styles manually, we first get different driving styles combining speeds and accelerations through a sticky HDP-HMM method automatically and then consider the mean of speeds and accelerations of each kind of driving styles to rank them.
As can be seen from Fig.\ref{fig: Ranking process}, we first divide the driving styles into four levels based on whether
$a_f$ is always greater/less than zero.
%%%%%%%%%%%%%%%%%%%%%%%%{}%%%%%%%%%%%%%%%%%%%%
%%%%%%%%%%%%%%%%%%%%%%%%%%%%%%%%%%%%%%%%%%%%
\begin{enumerate}[Level 1]
\item If $a_f$ is always greater than zero, it means the car is accelerating all the time,
which reflects that the driving scenario is very safe for the car. 

\item If the car accelerates at first and then slow down, it means the car will go into a dangerous status,
but there’re still a certain amount of response time. So it’s safe at first.

\item If the car slows down at first and then accelerate, it means the car is in a dangerous status but
then it gets out of the risky area. So it is a little dangerous.

\item If $a_f$ is always less than zero, it means the car is slowing down all the time,
which reflects that the driving scenario is very dangerous for the car.

\end{enumerate}

Then we analyze each level respectively. 

\begin{enumerate}[For Level 1:]
\item Because the car is accelerating all the time and to quantify the problem, we compare the mean of the velocity of each clustered driving style since it's also related to traffic accidents\cite{TANISHITA2017107}. 

\item Level 2 means that the car will accelerate at first but then slow down. 
So we mainly consider the first accelerating part and just like Level 1, we compare the mean of the velocity of each clustered driving style.

\item Level 3 means that the car will decelerate at first but then accelerate. 
We mainly consider the first decelerating part but firstly we need to estimate if the velocity will fall to zero in this part. If yes, it’s more risky and called Level 3.1. Otherwise, it’s less risky and called Level 3.2.
Then for Level 3.1, we compare the  average acceleration of the decelerating part. Because when a car is
 slowing down, the acceleration will mostly reflect the driver’s intents. It’s the same method with Level
  3.2.         

\item Finally we consider Level 4 which means that the car will decelerate all the time.
So just like the decelerating part in Level 3, we first estimate if the velocity will fall to zero in the 
decelerating process and divide them into Level 4.1 and Level 4.2 and then compare the mean $a_f$ of the whole decelerating
process for each level respectively.

\end{enumerate}

\subsection{Feature analysis and mapping} After clustering and ranking the urgency levels of driving styles, we map driving
scenarios to the ranked driving styles and analyze the relationship between driving styles and driving
scenarios according to the clustering and ranking results. 
Additionally, we use four variables to represent the features of driving scenarios:

\begin{itemize}

\item $Number$, the number of 3D bounding box in a scenario.
\item $Distance$, the distance between the nearest 3D bounding box and the ego car.
\item $Type$, the type of the nearest 3D bounding box like a cyclist or a car.
\item $Angle$, the observed angle between the ego car and the nearest 3D bounding box.

\end{itemize}

The information of 3D bounding box is also from the labels of KITTI dataset\cite{Geiger2012CVPR}.
By analyzing the internal cause that leads to
the clustering and ranking results, we can find how the driving scenarios affect the ego car's driving
styles and how the driving styles react to the driving scenarios at the same time. Finally, we get the 
conclusions in the experiment.

\section{EXPERIMENTS AND ANALYSIS}
\label{section: experiment and analysis}
This section will introduce our experiment to validate our methods. 

\subsection{Data Collection}
%%%%%%%%%%%%%%%%%%%%%%%%%%%%%%%%%%%%%%%%%%%%%%%%%
\begin{figure}
\centering
\includegraphics[width=3.5in]{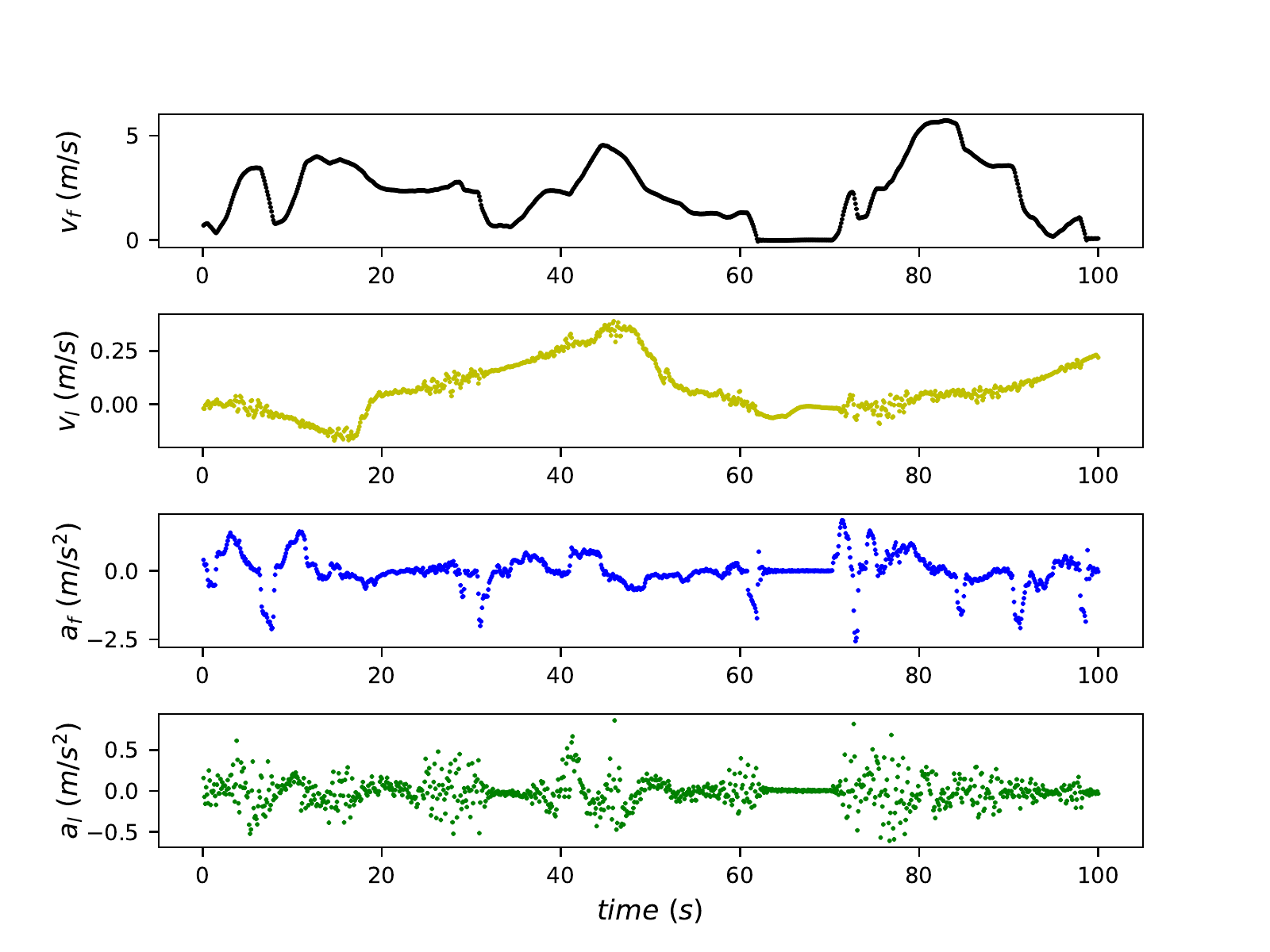}
\caption{An example of 100 seconds driving data including the change of $v_f, v_l, a_f \ and \ a_l$ over time.}
\label{fig: Data for clustering}
\end{figure}
%%%%%%%%%%%%%%%%%%%%%%%%%%%%%%%%%%%%%%%%%%%%%%%%%

%%%%%%%%%%%%%%%%%%%%%%%%%%%%%%%%%%%%%%%%%%%%%%%%%%
\begin{figure*}
\centering
\includegraphics[width=5in]{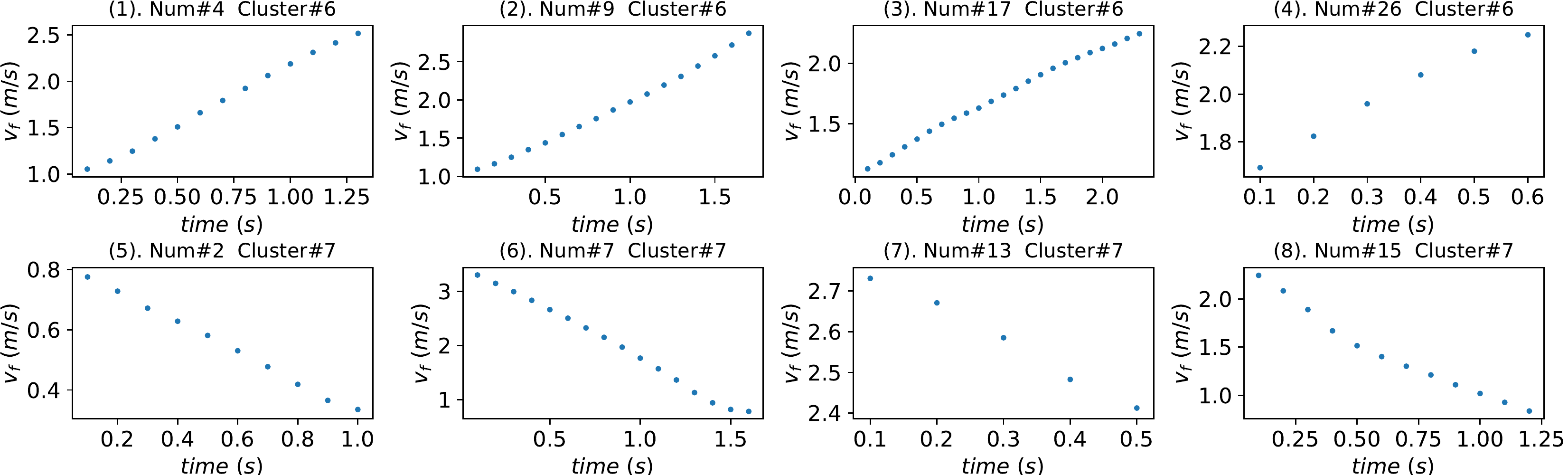}
\caption{Some examples of clustering results. Subgraph (1), (2), (3) and (4) belong
to Cluster\#6 while subgraph (5), (6), (7) and (8) belong
to Cluster\#7.}
\label{fig: sub examples of clustering}
\end{figure*}

%%%%%%%%%%%%%%%%%%%%%%%%%%%%%%%%%%%%%%%%%%%%%%%%%%
%%%%%%%%%%%%%%%%%%%%%%%%%%%%%%%%%%%%%%%%%%%%%%%%%%
\begin{figure}
\centering
\includegraphics[width=3.5in]{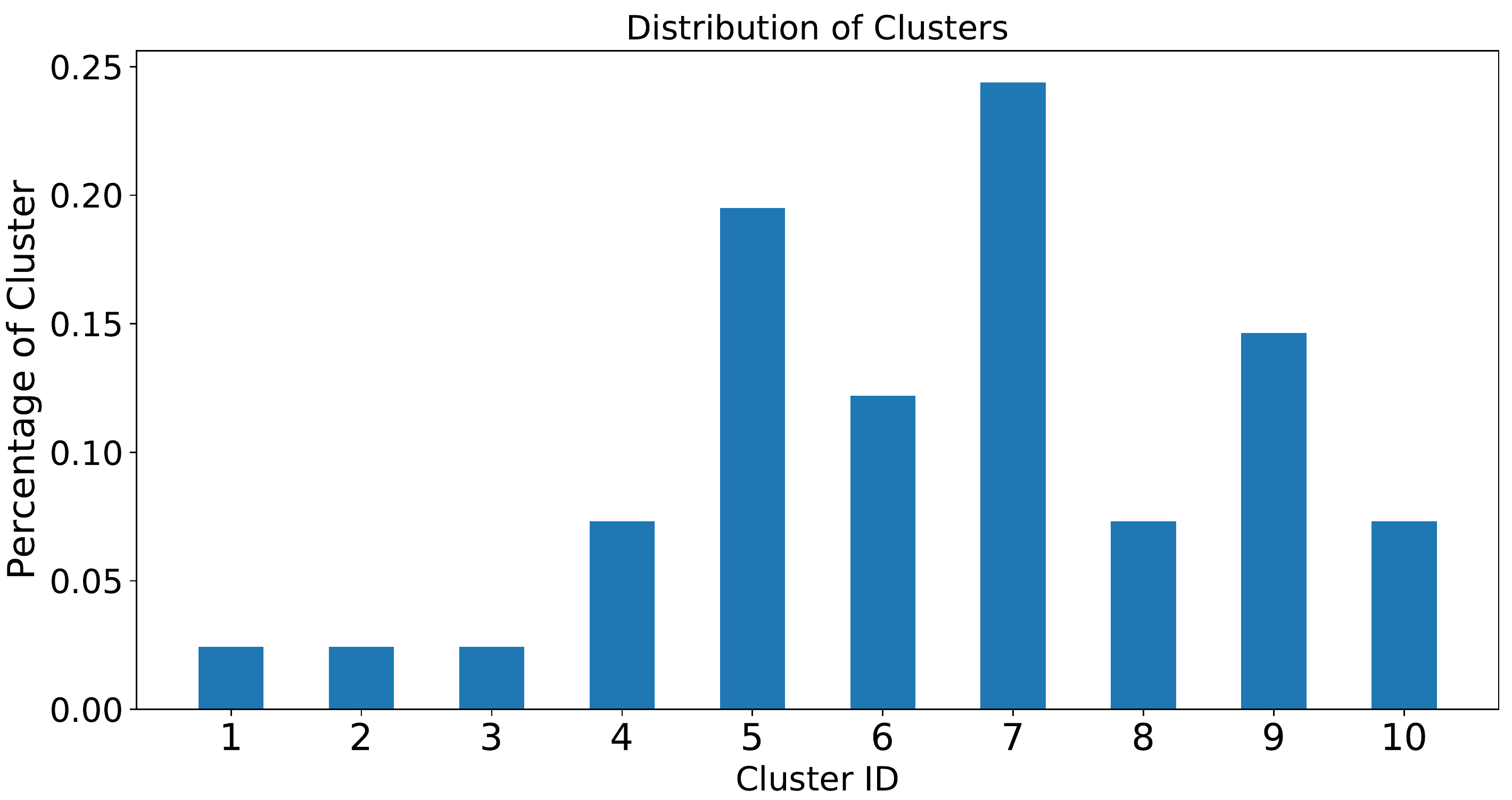}
\caption{This figure shows the distribution of clustering results using sticky HDP-HMM method.}
\label{fig: distribution of clustering}
\end{figure}
%%%%%%%%%%%%%%%%%%%%%%%%%%%%%%%%%%%%%%%%%%%%%%%%%%%
The driving data we use is from KITTI dataset\cite{Geiger2013IJRR} and the acquisition
frequency is 10Hz. To reflect driving styles more completely, we use acceleration and velocity to
discribe driving styles. Because acceleration can reflect a driver's intents directly and 
velocity can reflect a car's driving status more intuitively. More specifically, we consider 
four variables for use of clustering: forward velocity $v_f$, leftward velocity $v_l$, forward acceleration $a_f$
and leftward acceleration $a_l$.
Fig. \ref{fig: Data for clustering} shows an example of 100 seconds driving data including the change of 
$v_f$, $v_l$, $a_f$ and $a_l$ over time.

\subsection{Clustering}

As mentioned before, we use sticky HDP-HMM method to cluster the driving data which is shown as $[v_f^{1:T},v_l^{1:T},a_f^{1:T},a_l^{1:T}]$ and Pyhsmm \cite{johnson2013hdphsmm} is used to implement the parameters including $\alpha, \kappa, \gamma$, etc. to develop the sticky HDP-HMM method.

Fig.\ref{fig: distribution of clustering} shows the result where we can find that
the driving process is clustered into 10 clusters which means we get 10 kinds of driving styles. In addition, from the distribution of clusters, we find that, among the clusters, Cluster\#5 and \#7
account for a larger proportion while Cluster\#1, \#2 and \#3 account for a smaller 
proportion. So from the distribution we can infer that when a car is driving, situations like Cluster\#5 and \#7 happen more often, while situations like Cluster\#1, \#2 or \#3 don’t happen often.

Fig.\ref{fig: sub examples of clustering} shows some examples of the clustered results. From Fig.\ref{fig: sub examples of clustering} we know that subgraph
(1), (2), (3) and (4) belong to Cluster\#6 while subgraph (5), (6), (7) and (8) belong to Cluster\#7.
In addition, the figure shows the change of forward velocity $v_f$ over time. It's obvious that the 
forward velocity $v_f$ is increasing all the time in subgraph (1), (2), (3) and (4) belonging to Cluster\#6 
while the forward velocity $v_f$ is decreasing all the time in subgraph (5), (6), (7) and (8) belonging to  
Cluster\#7. So Cluster\#6 reflects the accelerating style of the car while Cluster\#7 shows the
decelerating style.

It also shows that the clustering results are reasonable since 
the driving styles in the same cluster are similar while the driving styles in the different clusters
are different.

%%%%%%%%%%%%%%%%%%%%%%%%%%%%%%%%%%%%%%%%%%%%%%%%%
\begin{figure}
\centering
\includegraphics[width=3.5in]{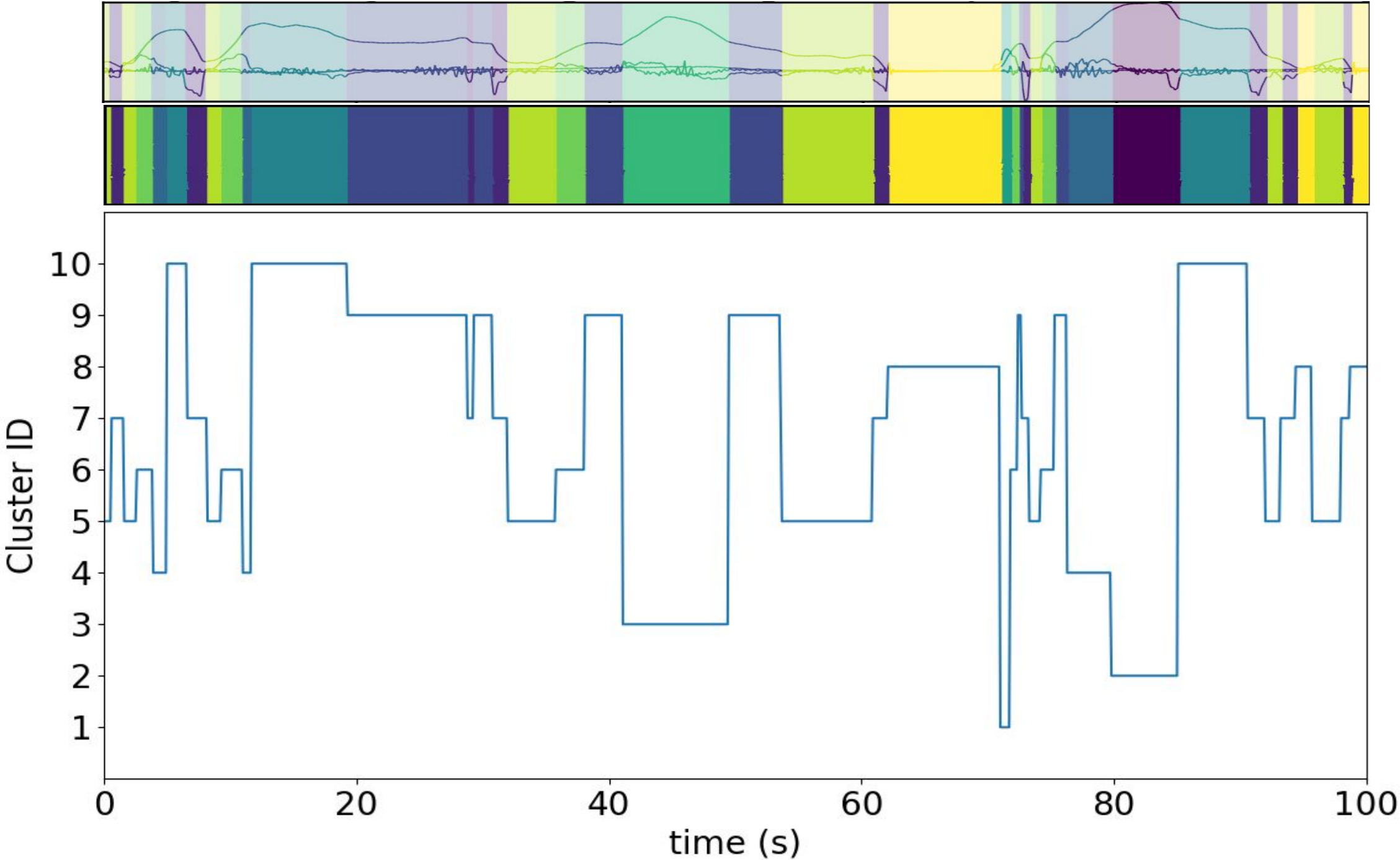}
\caption{In the color stripe, different colors represent different clusters and the whole color stripe shows the change of clusters over time which reflects the change of driving styles over time. In the line plot, the horizontal axis represents time
and the vertical axis represents the different clusters.}	
\label{fig: results of clustering combination}
\end{figure}
%%%%%%%%%%%%%%%%%%%%%%%%%%%%%%%%%%%%%%%%%%%%%%%%%%%

Fig.\ref{fig: results of clustering combination} shows the clustering results more specifically.
In the color stripe, different colors represent different clusters and the whole color stripe shows the change 
of clusters over time which reflects the change of driving styles over time. The line plot in 
Fig.\ref{fig: results of clustering combination} shows the change of different clusters over time in more 
details. The horizontal axis represents time and the vertical axis represents the different clusters. 
From this figure, we can learn about the change of vehicle driving styles with time during this period more 
clearly.

\subsection{Ranking}
\begin{figure}
\centering
\includegraphics[width=3in]{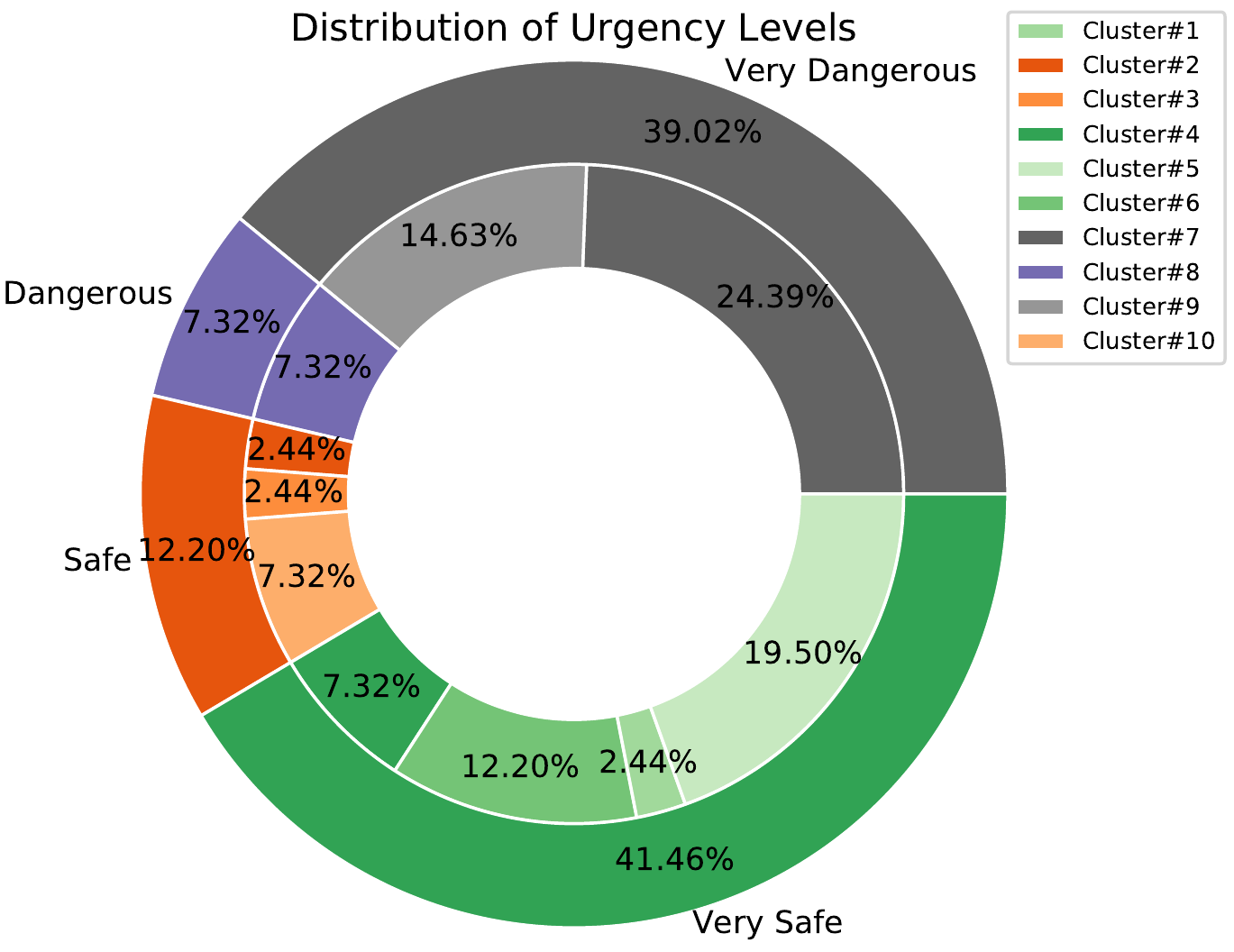}
\caption{The outer layer represents the distribution of clusters in the four levels and the layer inside stands for a more precise division of urgency levels. Besides, for the levels
belonging to the $Dangerous$ level or $Very \ Dangerous$ level, the darker the color, the higher the 
urgency level. Whereas for levels belonging to the $Safe$ or $Very \ Safe$ level, the stronger the color,
the safer it is. }
\label{fig: Ranking results}
\end{figure}
%%%%%%%%%%%%%%%%%%%%%%%%%%%%%%%%%%%%%%%%%%%%%%%%%%
Then we validate our ranking method.
We also use 100 seconds' driving data as an example. First, we divide the car's driving styles into
four different levels according to its acceleration. If the acceleration is always greater than zero,
it means the driving scenario is very safe for the car. Whereas always slowing down 
means the car is in a very dangerous state. Accelerating at first and then slowing down show that the car will go into a dangerous status, but there is still a certain amount of response time. On the contrary, if the car slows down at first and then accelerates, it means the car is in a dangerous status but then it gets out of the risky area. So we just call it dangerous.  After we get the four levels containing corresponding clusters,
we take their velocity and acceleration into account for each level. For instance, for clusters in 
Level\#1, we compare the average forward velocity in each cluster and divide the urgency levels for the 
clusters in Level\#1 more elaborately. More specific details are mentioned in Section \ref{section: methods}.

Then we get the result in Fig.\ref{fig: Ranking results}.
In Fig.\ref{fig: Ranking results}, the outer layer represents the distribution of clusters in the four levels.
Whereas the layer inside stands for a more precise division of urgency levels. In addition, for the levels
belonging to the $Dangerous$ level or $Very \ Dangerous$ level, the darker the color, the higher the 
urgency level. Whereas for levels belonging to the $Safe$ or $Very \ Safe$ level, the stronger the color,
the safer it is. 

So from Fig.\ref{fig: Ranking results} we can find that for the 10 clusters, Cluster\#1, \#4, \#5 and \#6 belong
to Level\#1($Very \ Safe$) and Cluster\#4 is safest. Besides, Cluster\#2, \#3, \#10 belong to 
Level\#2($Safe$) and Cluster\#2 is safest. Additionally, Cluster\#8 is in the $Dangerous$ level. And for
the Cluster\#7 and \#9 in the $Very \ Dangerous$ level, Cluster\#7 is more dangerous than Cluster\#9. Therefore we get urgency levels 
for different clusters and also their percentage distribution in Fig.\ref{fig: Ranking results}.

\subsection{Feature analysis and mapping}  
%%%%%%%%%%%%%%%%%%%%%%%%%%%%%%%%%%%%%%%%%%%%%%
\begin{figure}
\centering
\includegraphics[width=3in]{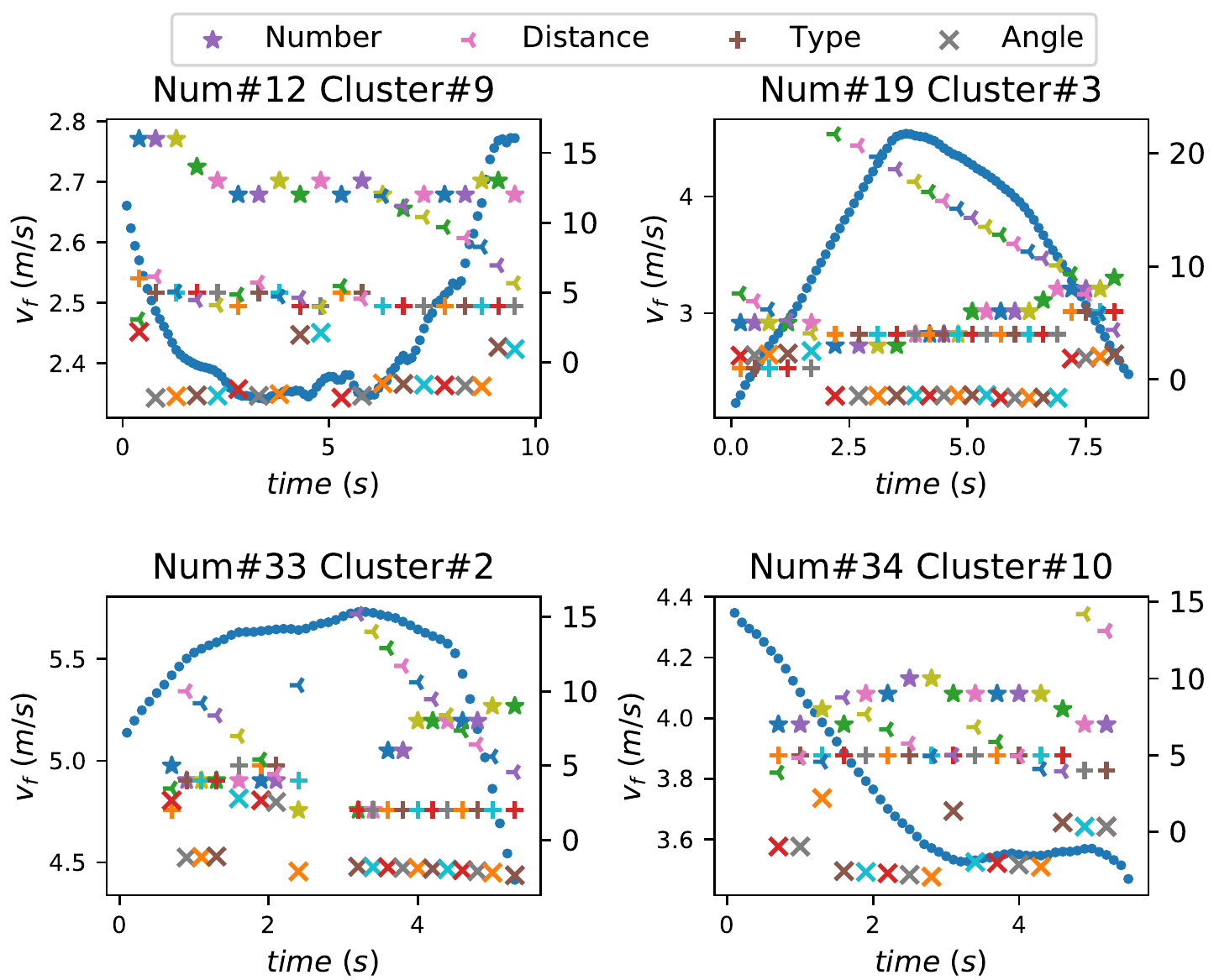}
\caption{This figure shows the change of five variables over time. The
vertical axis on the left represents the forward velocity
$v_f$ while the vertical axis on the right represents the four
variables $[Number,Distance,Type,Angle]$.}
\label{fig: Feature mapping}
\end{figure}
%%%%%%%%%%%%%%%%%%%%%%%%%%%%%%%%%%%%%%%%%%%%%
Finally, we map driving scenarios to the ranked driving styles and finally get the ranked driving scenarios of different risk levels. More importantly, we find the relationship between them. As mentioned in Section \ref{section: methods}, we use four variables: 
$[Number, Distance, Type, Angle]$ to represent the features of driving scenarios. Additionally, we use
another four variables $[v_f, v_l, a_f, a_l]$ to describe the driving styles. 
Thus, we use $[v_f, v_l, a_f, a_l]\Leftrightarrow[Number, Distance, Type, Angle]$

to describe the 
relationship between driving styles and driving scenarios. To explain the relationship between them
more concretely, we use the relationship $[v_f]\Leftrightarrow[Number, Distance, Type, Angle]$

as an example in Fig.\ref{fig: Feature mapping} which shows the change of the five variables over 
time. The vertical axis on the left represents the forward velocity $v_f$ 
while the vertical axis on the right represents the four variables $[Number, Distance, Type, Angle]$
of driving scenarios. 
From the analysis of the relationship between driving scenarios and driving styles, we get the conclusions
below:

\begin{enumerate}[(1)]
\item On the whole:
\begin{itemize}
\item $Type$ and $Angle$ do have an effect on the driving performance 
of the car, but less than the other two factors($Number$ and $Distance$).
\item In most cases, $v_f$ decreases as $Number$ increases and vice versa.
\item In most cases, $v_f$ decreases as $Distance$ decreases and vice versa.
\end{itemize}
\item Through a more detailed analysis of the relationship between driving scenarios and driving styles,
 we find:
\begin{itemize}
\item When the nearest object is close to the car, the change of $Distance$ is positively correlated with $v_f$. That means, the closer the object is to the car, the slower the car velocity is.
\item When the object is far away, even if $Distance$ decreases, $v_f$ may continue to increase or remain unchanged. It's reasonable. For instance, the driver may think it isn't so dangerous when it's far away. On the contrary, when the distance is too close, it's also possible for $v_f$ to increase. For example, there is no threat to the car after they pass each other.
\item As $Distance$ decreases, $v_f$ may decrease, but it may also be affected by $Number$ and other factors. Similarly, as $Number$ increases, $v_f$ decreases for the most part, but it is also affected by $Distance$ and some other factors.
\item When $Distance$ is reduced to a certain extent, it may suddenly jumps to a larger value. Probably because the nearest
object has changed a new one. Then we need to change our focus in time.

\end{itemize}
\end{enumerate}

For instance, in Fig.\ref{fig: Feature mapping}, for the subgraph named $Num\#12 \ Cluster\#9$, with the 
decreasement of $Distance$ at first, $v_f$ also decreases. However, later the value of $Distance$
suddenly jumps to a larger value. This reflects that the first nearest object has passed the ego
car and now the original second nearest object becomes the nearest object to the ego car at this time. 
In addition, we can also find that $v_f$ decreases as the $Number$ increases and the $Distance$ decreases in $Num\#33 \ Cluster\#2$.

\section{CONCLUSION AND FUTURE WORK}
\label{section: conclusion and future work}
In this paper, we propose a framework to estimate risk levels of driving scenarios and analyze
the relationship between driving scenarios and driving styles. There're four steps in our
framework: clustering driving styles, ranking driving styles, feature analysis and mapping. We
use sticky HDP-HMM to cluster driving styles and get reasonable results. After ranking driving styles and
mapping we can get different risk levels of driving scenarios. Besides, through 
analyzing the relationship between driving styles and driving scenarios, we get reasonable 
conclusions about the relationship between them and know how the different driving
scenarios will affect a car's driving status to different extent.

There're still future work we need to do though. In our ranking system, we mainly consider forward acceleration and velocity. In our future work, we'll also take leftward and upward velocity and acceleration into account. In fact, after we get different clusters of driving styles and get the
relationship between driving scenarios and driving styles, we can classify a new driving scenario
into a risk level and then help the car to make more precise and safer decisions.

%%%%%%%%%%%%%%%%%%%%%%%%%%%%%%%%%%%%%%%%%%%%%%%%%%%%%%%%%%%%%%%%%%%%%%%%%%%%%%%%

\section*{ACKNOWLEDGMENT}
This work was finished when Songlin Xu was a short-term scholar under the supervision of Prof. Ding Zhao at Carnegie Mellon University. He would like to thank Prof. Ding Zhao for his suggestions for this work.


\bibliographystyle{IEEEtran} 
% \bibliography{IEEEabrv,root}

\end{document}